\def\year{2020}\relax
\documentclass[letterpaper]{article} 
\usepackage{aaai20}  
\usepackage{times}  
\usepackage{helvet} 
\usepackage{courier}  
\usepackage[hyphens]{url}  
\usepackage{graphicx} 
\usepackage{multirow}

\usepackage{epsfig}
\usepackage{amssymb}
\usepackage{amsmath,bm}
\usepackage{sidecap}
\sidecaptionvpos{figure}{t}

\newcommand*{\pd}[3][]{\ensuremath{\frac{\partial^{#1} #2}{\partial #3}}}
\usepackage{pifont}
\newcommand{\etal}{\emph{et al.}}

\title{Fine-grained Recognition: Accounting for Subtle\\ Differences between Similar Classes}
\author{Guolei Sun,\textsuperscript{\rm 1} Hisham Cholakkal,\textsuperscript{\rm 2} Salman Khan,\textsuperscript{\rm 2} Fahad Shahbaz Khan,\textsuperscript{\rm 2} Ling Shao\textsuperscript{\rm 2}   \\ 
\textsuperscript{\rm 1}ETH Zurich, \textsuperscript{\rm 2}Inception Institute of Artificial Intelligence\\ 
guolei.sun@vision.ee.ethz.ch, \{hisham.cholakkal, salman.khan, fahad.khan, ling.shao\}@inceptioniai.org 
}
\begin{document}

\maketitle

\begin{abstract}
The main requisite for fine-grained recognition task is to focus on subtle discriminative details that make the subordinate classes different from each other. We note that existing methods implicitly address this requirement and leave it to a data-driven pipeline to figure out what makes a subordinate class different from the others. This results in two major limitations: \emph{First,} the network focuses on the most obvious distinctions between classes and overlooks more subtle inter-class variations. \emph{Second,} the chance of misclassifying a given sample in any of the negative classes is considered equal, while in fact, confusions generally occur among only the most similar classes. Here, we propose to explicitly force the network to find the subtle differences among closely related classes. In this pursuit, we introduce two key novelties that can be easily plugged into existing end-to-end deep learning pipelines. On one hand, we introduce ``diversification block" which masks the most salient features for an input to force the network to use more subtle cues for its correct classification. Concurrently, we introduce a ``gradient-boosting" loss function that focuses only on the confusing classes for each sample and therefore moves swiftly along the direction on the loss surface that seeks to resolve these ambiguities. The synergy between these two blocks helps the network to learn more effective feature representations. Comprehensive experiments are performed on five
challenging datasets. Our approach outperforms existing methods using similar experimental setting on all five datasets.
\end{abstract}

\begin{figure}[ht]
\centering
\includegraphics[width=0.86\linewidth]{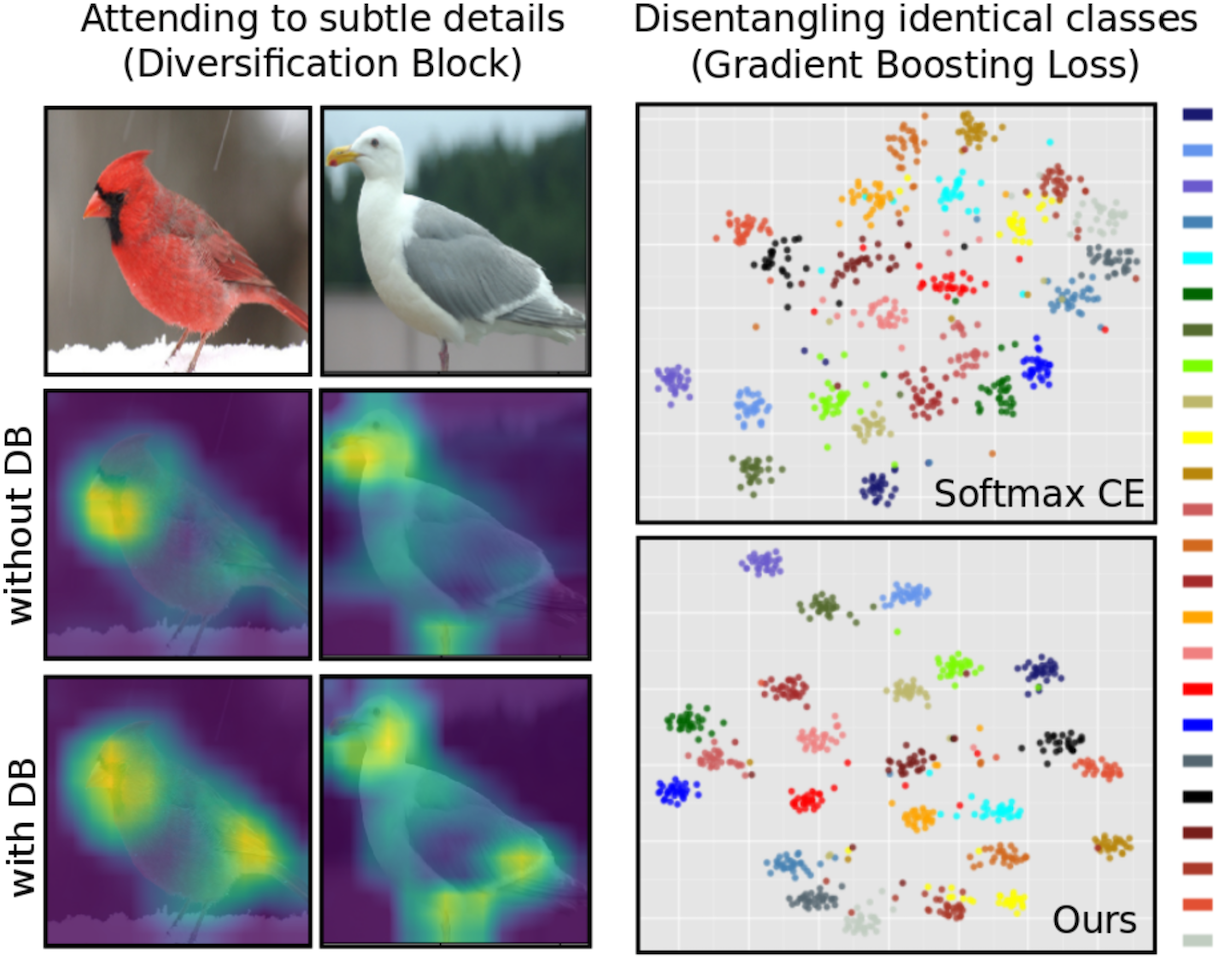}\\
\caption{Illustration of two novel components of our approach. \emph{Left:} comparison between class activation maps obtained from the model with our diversification block (DB) and the one without DB. Our DB forces the network to capture more discriminative regions. With DB (\emph{below}), the network finds beak, tail and feet of the bird as informative regions, while without DB (\emph{middle}), the network only focuses on beak. \emph{Right:} visual comparison in terms of $2$-$d$ tSNE \cite{van2014accelerating} plot for features of 24 kinds of \emph{Walbler} (confusing and difficult classes)  in CUB-200-2011 between network trained with cross entropy (CE) (\emph{top}) and our gradient-boosting loss (\emph{below}). By focusing on difficult classes, our gradient-boosting loss can distinguish between hard classes which are not well separated by CE. 
}
\label{Fig:intro_image}
\end{figure}

\section{Introduction}
Fine-grained recognition focuses on discriminating between children classes of a main parent category (e.g., cars \cite{dataset_cars}, dogs \cite{dataset_dogs}, birds  \cite{dataset_cub}, and aircrafts \cite{dataset_aircraft}). Deep CNNs have excelled immensely on traditional visual recognition tasks where categories greater differ from each other.  However, fine-grained visual categorization (FGVC) poses a significant challenge mainly due to the close resemblance between subcategories e.g., different species of the same bird. The challenge is compounded by the fact that the classifier has to be invariant to intra-class variations, e.g., pose, appearance and lighting changes.

Common deep learning based approaches for FGVC learn a mapping between input images and output labels. While doing so, a natural tendency during learning is to focus on only few distinguishing parts in an object to deal with confusing inter-class similarities and large intra-class variations (see Fig. \ref{Fig:intro_image}). The analysis of attention based models provides the evidence that attention maps are often densely concentrated on a few parts, thus considering only a limited set of cues. In contrast, here we propose to spread the attention to  consolidate a diverse set of relevant cues spread across the activation map. While we diversify attention at the feature level, we do the opposite at the prediction level, i.e., focus on only the most confusing cases to achieve better discriminability. Popular loss functions such as cross entropy, consider all classes to compute the error signal for parameter update. When closely related classes are present in the data, this leads to a weak supervision signal resulting in slower convergence and low recall rates. We show that selectively attending to the hard negative 
classes helps in achieving much faster convergence and higher accuracy.

Our approach can also be understood as a mechanism to enhance network generalization and avoid overfitting. This consideration is of particular relevance to FGVC, since the datasets are generally smaller due to the high cost of obtaining fine-grained annotations from experts. In effort to minimize loss on training data, a high-capacity network can end up associating unrelated concepts (such as those of background) to the fine-grained object itself. By concentrating on only the closely related classes and diversifying the model's attention, we are in fact regularizing the model to avoid overfitting the training samples. Our approach reduces classifier’s confidence on training samples and therefore makes it more generalizable. We note that regularization schemes such as label-smoothing \cite{szegedy2016rethinking} and maximum prediction entropy \cite{max_entropy} are related to ours, but significantly different as we impose regularization on both features and output predictions.

Our main contributions are as follows:
\begin{itemize}\setlength{\itemsep}{0em}
\item We introduce a \emph{gradient-boosting} loss  that seeks to resolve ambiguities among closely related classes by appropriately magnifying the gradient updates.
\item Our \emph{diversification block} masks out the salient features in order to force the network to look for subtle differences between similar-looking categories.
\item The proposed method makes the convergence faster while outperforming existing methods on five datasets.
\end{itemize}

\section{Related Works}
Fine-grained classification has attracted much research attention in the recent years. 
Despite several attempts \cite{nts_2018,mamc_2018}, 
FGVC is still an active research problem. 
To deal with the problem of subtle intra-class distance, many approaches focused on obtaining more relevant features~\cite{fine_grain_annotation1,lin2015bilinear,cbp,nts_2018,mamc_2018}. One of the earliest but naive strategy was to exploit part annotations \cite{fine_grain_annotation1} to locate the objects so that more informative features were used.  Such an approach requires more labeling effort and has therefore limited scalability. Another stream of works ~\cite{lin2015bilinear,cbp,Li_2018_CVPR} developed complex pooling methods, so that complex local features can be used for classification. However, one obvious drawback of those methods is the high computation complexity. To deal with the problem of small fine-grained datasets, Cui {\etal} ~\cite{cui2018large} proposed a transfer learning scheme from selected subset of the source domain to target domain. However, it requires to re-train models on a subset of large datasets like ImageNet \cite{ILSVRC15} and iNaturalist \cite{van2018inaturalist}.

Recent efforts~\cite{nts_2018,mamc_2018,Chen_2019_CVPR,zheng2019looking,ge2019weakly} used only class labels to automatically locate informative regions. 
Specifically, Yang \etal ~\cite{nts_2018} adapted a Navigator-Teacher-Scrutinizer system under a multi-stage scheme. Sun \etal ~\cite{mamc_2018} leveraged multiple channel attentions to learn several relevant regions. Wang \etal ~\cite{DFL_cnn_2018} used a bank of convolutional filters to capture discriminative regions in the feature maps.
Chen ~\cite{Chen_2019_CVPR} deconstructed and reconstructed input images to find discriminative regions and features. Zheng \etal ~\cite{zheng2019looking} proposed trilinear attention sampling network to learn features from different details. 
Despite the fact that the above methods perform well, they generally need to be trained in multiple stages or learn high-dimension features, resulting in increased training times. Another recent work \cite{ge2019weakly} developed a computationally complex, three-stage pipeline for fine-grained classification. Their framework requires a weakly supervised object detector, a mask-rcnn \cite{maskrcnn} based instance segmentation and an LSTM for capturing the context. Moreover, the mask-rcnn needs to be pretrained on an additional dataset: MS-COCO \cite{coco_eccv2014}. 
Our proposed diversification block adopts a novel way to find more relevant features by suppressing the most prominent discriminative regions in class activation maps  \cite{CAMs} and thus forcing the network to find other informative regions. We note that hide-and-seek \cite{has_2017} is related to ours, but largely different since our module works on feature maps and selectively suppresses discriminative regions. 
Our module is trained end-to-end with a computational cost nearly equal to the backbone.

Lately, FGVC strategies aimed to learn optimal classifiers on top of deep features have been proposed \cite{pc_2018,max_entropy}. Qian \etal~\cite{qian2015fine} employed a multi-stage framework which accepted pre-computed feature maps and learned the distance metric for classification.   Dubey \etal~ \cite{pc_2018} adapted the idea from pairwise learning and used Siamese-like neural network. 
A triplet loss was used in \cite{wang2016mining} to achieve better inter-class separation. The contrastive and triplet losses, however, increase the computational cost of training. \cite{max_entropy} proposed a maximum entropy loss for fine-grained classification by using the principle of maximum-entropy. All above methods do not specifically focus on differentiating confusing classes. Further, all the negative categories for a given sample are considered as equal. Our proposed gradient-boosting loss solves the problem by explicitly focusing on hard classes, incurs no additional cost and provides faster convergence rates.

\begin{figure*}[t]
\centering
	\includegraphics[width=0.98\linewidth]{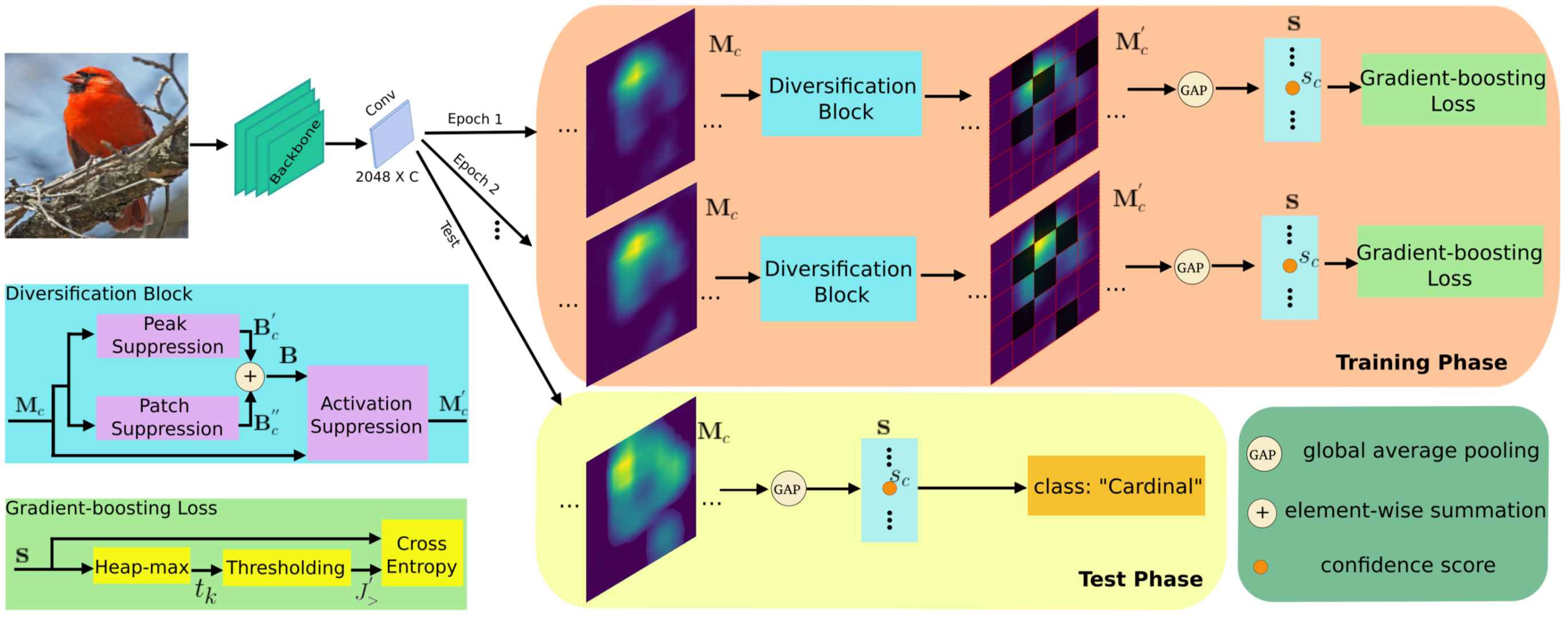}\\
\caption{Overview of our overall architecture. Our method contains two novel components: diversification block and gradient-boosting loss. The diversification block suppresses the discriminative regions of the class activation maps, and hence the network is forced to find alternative informative features. The gradient-booting loss focuses on difficult (confusing) classes for each image and boosts their gradient. As a result, the network moves swiftly (faster convergence) to discriminate the hard classes.}
			\label{Fig:architectue}
\end{figure*}

\section{Method}
In this section, we introduce our method which can be easily plugged into any classification network. As shown in Fig. \ref{Fig:architectue}, to deploy our approach, we need to replace global pooling layer and the last fully connected layer of the backbone network with a 1$\times$1 convolution having output channels equal to the number of classes. Our method includes two novel components: (a) A diversification module which forces the network to capture more subtle features, rather than only the most obvious ones; (b) A gradient boosting loss  which trains the network to focus on highly confusing classes. These two components will be addressed in this section.

\subsection{Diversification Block}\label{part_enhan}
Consider the multi-class image classification task with $C$ classes as shown in Fig. \ref{Fig:architectue}. Let $I$ be a training image with ground-truth label $l \in J$, where $J= \{1,2,...,C\}$ is the label set containing all labels. The input to our diversification block is the category-specific activation maps $\mathbf{M} \in R^{C \times H \times W}$, which is the output of the modified network. We denote $ \mathbf{M}=\{\mathbf{M}_c : c\in [1,C]\}$, where $\mathbf{M}_c \in \mathbb{R}^{H \times W}$ is the individual activation map corresponding to $c^{th}$ class.  Here, $H$ and $W$ refer to the height and width of the output activation maps.  

The basic idea of our diversification block is to suppress the discriminative regions of the activation map $\mathbf{M}$, so that the network is forced to look for other informative regions which is expected to enhance classification performance. In the following, we will target two relevant questions: (1) Where to suppress information? and (2) How to suppress?

\subsubsection{Mask Generation} \label{sup_loca}
Here, we explain the procedure to generate the mask that indicates the locations in $\mathbf{M}$ that are suppressed. Let $\mathbf{B} = \{\mathbf{B}_c : c \in [1,C]\}$, where $\mathbf{B}_c \in \mathbb{R}^{H\times W}$ denotes the binary suppressing mask for its corresponding activation map $\mathbf{M}_c$. Each element in mask $\mathbf{B}_c$ is in the domain $\{0,1\}$, where $1$ indicates the corresponding location will be suppressed while $0$ means that no suppression will take place.

\textbf{\textit{Peak suppression:}} First, we randomly suppress the peak locations of the activation maps because they are the most discriminative regions for the classifier. By suppressing the peaks, the network is forced to find alternative relevant regions in the image. Let ${\mathbf{P}}_{c}  \in \mathbb{R}^{H\times W}$ be the peak map derived from $c^{th}$ object category map ($\mathbf{M}_{c}$) such that:
\begin{equation}
\small
 {\mathbf{P}}_{c}(i,j)=\begin{cases}
    1,&\text{if $\mathbf{M}_{c}(i,j)=\max(\mathbf{M}_{c}$}),\\
    0, & \text{otherwise}.
  \end{cases}
\end{equation}
Here, $\max(\mathbf{M}_{c})$ denotes the maximum of matrix $\mathbf{M}_{c}$. We suppress the peaks of different object categories with probability $p_{peak}$. The masks $\mathbf{B}^{'}_{c}$ to randomly hide the peaks are generated as follows:
\begin{equation}
    \mathbf{B}^{'}_{c} = r_c * {\mathbf{P}}_{c}, \quad \text{where} \quad r_c \sim \text{Bernoulli}(p_{peak}),
\end{equation}
where '$*$' denotes element-wise multiplication and $r_c$ is a Bernoulli random variable that has $p_{peak}$ probability of being 1.

\textbf{\textit{Patch suppression:}}
Peaks are the most discriminative regions, but there are other discriminative regions as well that encompass more subtle inter-class differences. In the following, we explain how to suppress locations other than peaks in the activation maps. We divide each $\mathbf{M}_{c}$ into a grid of patches, where each fixed sized patch $\mathbf{M}_c^{[l,m]} \in \mathbb{R}^{G\times G}$ is indexed by row $l$ and column $m$. Lets assume the set of all such patches on the grid is given by:
\begin{align}\label{eq:patches}
    \mathbf{G}_c = \{\mathbf{M}_c^{[l,m]} : l \in [1,\frac{W}{G}], m \in [1,\frac{H}{G}]\}.
\end{align}
After this operation, the activation map $\mathbf{M}_{c}$ will be divided into $(W\times H)/G^2$ patches. Let $\mathbf{B}^{''}_{c} \in \mathbb{R}^{H\times W}$ be the mask for randomly hiding patches for $c^{th}$ activation map $\mathbf{M}_{c}$.  For each patch inside $\mathbf{M}_{c}$, we randomly hide it with probability $p_{patch}$ and set the elements of corresponding locations of $\mathbf{B}^{''}_{c} $ as 1. Otherwise, the elements of $\mathbf{B}^{''}_{c}$ are set to $0$:
\begin{align}
    \mathbf{B}^{''}_{c} = \{ \mathbf{B}^{''[l,m]}_{c} \in [\mathbf{0}, \mathbf{1}] \sim \text{Bernoulli}(p_{patch})\},
\end{align}
where, $\mathbf{0},\mathbf{1}\in \mathbb{R}^{G\times G}$ and $l,m$ are in the same range as Eq.~\ref{eq:patches}. To consider only the non-peak locations, we then set the element of $\mathbf{B}^{''}_{c}$ in the peak location of $\mathbf{M}_{c}$ as 0,
\begin{equation}
    \mathbf{B}^{''}_{c}(i,j)=0, \text{ if $\mathbf{M}_{c}(i,j)=\max(\mathbf{M}_{c}$)}.
\end{equation}
The final suppressing mask for $c^{th}$ category is obtained as:
\begin{equation}
    \mathbf{B}_c=\mathbf{B}^{'}_{c}+\mathbf{B}^{''}_{c}.
\end{equation}



\subsubsection{Activation Suppression Factor}\label{sup_values}
Setting values that replace the suppressed features is of much importance to achieve good performance. Let $\mathbf{M}^{'}=\{\mathbf{M}^{'}_c : c\in [1,C]\}$ represents the category activation maps obtained after our diversification module, which is generated as follows.
\begin{equation}
    \mathbf{M}^{'}_c(i,j)=\begin{cases}
    \mathbf{M}_c(i,j), \text{ if $\mathbf{B}_c(i,j)=0$},
    \\
    \alpha* \mathbf{M}_c(i,j), \text{ if $\mathbf{B}_c(i,j)=1$},
    \end{cases}
\end{equation}
where, $\alpha$ denotes the suppressing factor. Basically, we replace the values in the suppressing locations as $\alpha$ times of their initial values. In general, setting $\alpha$ to a low number will lead to good performance. Throughout our experiments, we set $\alpha$ as 0.1.

After feature masking, we perform global average pooling to get the confidence scores $\mathbf{s} \in \mathbb{R}^{1\times C}$  as follows:
\begin{equation}
 \mathbf{s} =\{s_c : c\in [1,C]\},\quad s_c=\text{AvgPool}(\mathbf{M}^{'}_c),
\end{equation}
where, $\text{AvgPool}$ denotes global average pooling.

\subsection{Gradient-boosting Cross Entropy Loss}\label{selected_loss}
While diversification module aims at finding more subtle variations in the input images, our second contribution is  a gradient-boosting loss function that specifically focuses on confusing classes to avoid misclassifications between them. We elaborate the proposed loss function below. 

\subsubsection{Loss Function}
The most widely used loss for image classification is cross entropy (CE) loss. For an image $I$, CE loss can be written as follows:
\begin{equation}\label{euq1}
    \text{CE}(\mathbf{s},l)=-\log \frac{\exp{(s_{l})}}{\sum_{i \in J}\exp{(s_i)}},
\end{equation}
where $l$ is the ground-truth label for image $I$. Here, the loss considers all negative classes equally. However, in fine-grained classification, the ground-truth class is generally much closer to a related subset of classes than others. For example, in CUB-200-2011 \cite{dataset_cub}, bird class of \emph{Acadian Flycatcher} is more closer to categories such as \emph{Great Crested Flycatcher}, \emph{Least Flycatcher}, \emph{Olive sided Flycatcher} and other kinds of \emph{Flycatcher}, since they all belong to the same species. As a result, the network is prone to making mistakes among these similar (thus confusing) classes and predicting relatively higher confidence scores for them. Based on this observation, we argue that the loss should focus more on the confusing classes, rather than simply considering all negative classes equally for the normalization in Eq.~\ref{euq1}. Hence, we propose a novel and simple gradient-boosting cross entropy (GCE) loss which focuses only on $k$ negative classes with top-$k$ highest confidence scores among all negative classes. Here, $k$ simply means the number of negative classes to focus on. We will show in the next section, that the proposed loss basically boosts gradients to more swiftly resolve ambiguities between closely related confusing classes.  

We define $J^{'}$ as the set of all negative classes, where $J^{'}=\{i : i \in [1,C] \wedge i \neq l\}$. Let $\mathbf{s}'=\{s_i,i \in J^{'} \}$ be the set containing confidence scores of all negative classes. We get the $k^{th}$ highest values of $\mathbf{s}'$ by heap-max algorithm \cite{max_heap} and denote it as $t_k$. Next, we split $J^{'}$ into $J^{'}_{>}$ and $J^{'}_{<}$ by thresholding $\mathbf{s}$ using $t_k$, defined as follows:
\begin{equation}\label{equ_set1}
    J^{'}_{>}=\{i: i \in J \wedge s_i \geq t_k\}
\end{equation}
\begin{equation}\label{equ_set2}
    J^{'}_{<}=\{i: i \in J \wedge s_i<t_k\},
\end{equation}
where, $J^{'}_{>}$ contains the negative classes whose confidence scores are within the top-$k$ of all negative classes, and $J^{'}_{<}$ is the set of negative classes whose confidence scores rank below the top-$k$ classification scores.  

Instead of considering all negative classes in Eq.~ \ref{euq1}, our gradient-boosting cross entropy loss only focuses on confusing classes ($J^{'}_{>}$). The negative classes in $J^{'}_{<}$ do not contribute to the loss since the network can easily distinguish them from the ground-truth class. Our proposed loss is given by:
\begin{equation}\label{gce_loss}
    \text{GCE}(\mathbf{s},l)=-\log \frac{\exp{(s_{l})}}{\exp{(s_{l})}+\sum_{i\in J^{'}_{>}}\exp{(s_i)}}.
\end{equation}

As shown in Eq.~\ref{equ_set1} and \ref{gce_loss}, GCE loss focuses only on $J^{'}_{>}$, containing $k$ negative classes with top-$k$ highest confidence scores. Here, $k$ is a hyper-parameter (we found $k=15$ works best in our experiments). When $k=C$, GCE is equivalent to CE. 

In the following analysis, we will show how our loss can boost the gradient for both the ground-truth class and confusing negative classes.


\subsubsection{Gradient Boosting}
We analyze the loss from the perspective of gradient. For the original cross entropy (CE) loss, the gradient for $s_c$ is computed as:
\begin{equation}
    \pd{\text{CE}(\textbf{s},l)}{s_c}=\begin{cases}
    \frac{\exp{(s_c)}}{\sum_{i\in J}\exp{(s_i)}}, & c\neq l\\
    \frac{\exp{(s_c)}}{\sum_{i\in J}\exp{(s_i)}}-1, & c= l 
    \end{cases}
\end{equation}
For our gradient-boosting cross entropy loss, the gradient for $s_c$ is computed as:
\begin{equation}
    \pd{\text{GCE}(\textbf{s},l)}{s_c}=\begin{cases}
    \frac{\exp{(s_c)}}{\exp{(s_{l})}+\sum_{i\in J^{'}_{>}}\exp{(s_i)}}, & c\in J^{'}_{>}\\
    \frac{\exp{(s_c)}}{\exp{(s_{l})}+\sum_{i\in J^{'}_{>}}\exp{(s_i)}}-1, &c= l
    \end{cases}
\end{equation}
From our definition in Eq. \ref{equ_set1} and Eq. \ref{equ_set2}, the following relation exists between $J^{'}_{>}$ and $J^{'}$,
\begin{equation}
    J^{'}_{>} + \{l\} \subset J^{'}+\{l\}= J.
\end{equation}
As such, we obtain,
\begin{equation}
    \pd{\text{GCE}(\mathbf{s},l)}{s_c}> \pd{\text{CE}(\mathbf{s},l)}{s_c}.
\end{equation}

We can see that for both the ground-truth class and confusing negative classes, the gradient of our proposed loss is larger than the gradient of the original cross entropy loss. With our novel loss, the network can focus on differentiating difficult classes from the ground-truth class and converge faster, which is validated by our experiments.

\subsection{Training and Inference}
Our method is trained end-to-end in a single stage. The diversification block is only used during the training phase. As shown in Fig.~\ref{Fig:architectue}, during the training phase, class activation maps are passed through our novel diversification block and then to the global average pooling. As a result, discriminative regions are randomly masked and the network is forced to find other relevant areas. During test phase, the whole class activation maps are passed to global average pooling directly, without being suppressed at any region so that all informative regions found during training phase contribute to the final confidence score.

\section{Experiments}
\subsection{Datasets}
We comprehensively evaluate our algorithm on CUB-200-2011 \cite{dataset_cub}, Stanford Cars \cite{dataset_cars}, FGVC Aircraft \cite{dataset_aircraft}, and Stanford Dogs \cite{dataset_dogs}, all of which are widely used for fine-grained recognition. Statistics of all datasets are shown in Table~\ref{tab:dataset}. We follow the same train/test splits as in the table. For evaluation metric, we use top-1 accuracy following \cite{mamc_2018,pc_2018,max_entropy}.

Furthermore, we also evaluate on the recent terrain dataset for terrain recognition: GTOS-mobile \cite{dataset_gtosmobile} dataset and GTOS (Ground Terrain in Outdoor Scenes) \cite{dataset_gtos} dataset, which have potential use for autonomous agents (automatic car). The datasets are large-scale, containing classes of outdoor ground terrain, i.e. glass, sand, soil, stone-cement, and so on. Since those terrain classes are closely related, visually similar and thus difficult to classify, we use this challenging dataset to evaluate our method. Following \cite{dataset_gtosmobile}, we use GTOS as training and GTOS-mobile as test.
\begin{table}[h]
\centering
\begin{tabular}{|c|c|c|c|}
\hline
Dataset        & \#Class & \#Train & \#Test \\ \hline\hline
CUB-200-2011  & 200     & 5,994    & 5,794   \\ \hline
Stanford Cars  & 196     & 8,144    & 8,041   \\ \hline
FGVC Aircraft  & 100     & 6,667    & 3,333   \\ \hline
Stanford Dogs  & 120     & 12,000    & 8,580   \\ \hline
GTOS-mobile & 31     & 31,315    & 6,066   \\ \hline
\end{tabular}
\caption{Five commonly used benchmarks.}
\label{tab:dataset}
\end{table}

\subsection{Implementation Details}
For fair comparisons with other methods \cite{nts_2018,DFL_cnn_2018}, we use an input image resolution of 448$\times$448 in all experiments. 
We fine-tune pretrained network (ResNet-50 \cite{resnet_2016}) using our proposed diversification block and gradient-boosting loss due to its popularity in existing works. Momentum SGD optimizer is used with an initial learning rate of 0.001, which decays by 0.1 for every 50 epochs. We set weight decay as $10^{-4}$. Our algorithm is implemented using Pytorch \cite{paszke2017automatic} using two Tesla V100 GPU.

\subsection{Quantitative Results}
Our method does not require any part-annotation and can be trained using only class labels. Moreover, it is parameter-free and does not  increase the number of parameters compared to the ResNet-50 backbone. Our results are compared with the most recent and top-performing approaches evaluated under similar experimental setting. Several  approaches such as RACNN \cite{racnn_2017}, RAM \cite{li2017dynamic}, and NTS-net \cite{nts_2018} extract multiple crops at different scale from an input image. The classification score obtained from these crops are averaged to predict the final class during inference. For fair comparison, we report our  `multi-scale'  (five crops)  results, in addition to the  `single-scale' using one crop from an image.

The comparisons with various methods on four challenging fine-grained datasets, namely CUB-200-2011 \cite{dataset_cub}, FGVC Aircraft \cite{dataset_aircraft}, Stanford Cars \cite{dataset_cars}, and Stanford Dogs \cite{dataset_dogs},  are shown in Table \ref{tab:cub}. Additionally, results for GTOS-mobile \cite{dataset_gtos} are shown in Table \ref{tab:gtos}.  Overall, our proposed method outperforms previous methods on \emph{all} five datasets.

\begin{table*}[h]
\centering
\resizebox{2.1\columnwidth}{!}{
\begin{tabular}{|c|c|c|c|c|c|c|c|}
\hline
\multirow{2}{*}{Methods} & \multirow{2}{*}{Backbone} & \multirow{2}{*}{Resolution} & \multirow{2}{*}{\#Parameters} & \multicolumn{4}{c|}{Accuracy}          \\ \cline{5-8} 
&      &     &      & \textbf{CUB-200-2011} & \textbf{Aircrafts} & \textbf{~~Cars~~} & \textbf{~~Dogs~~}
 \\ \hline \hline
RACNN \cite{racnn_2017} & VGG-19 & 448 & 429M & 85.3 &88.2&92.5&87.3\\\hline
RAM  \cite{li2017dynamic} & ResNet-50  & 448 & $>$23.9M & 86.0&-& 93.1 &-\\\hline
MACNN \cite{zheng2017learning} & VGG-19 & 448 &144M & 86.5 &89.9&92.8&-\\\hline
MAMC \cite{mamc_2018} & ResNet-50 & 448   & 434M  & 86.3 &-& 93.0&85.2\\ \hline
MaxEnt \cite{max_entropy} & ResNet-50 & -  & 23.9M   & 86.5 &89.8&93.9&83.6\\ \hline
PC \cite{pc_2018} & ResNet-50 & -   & 23.9M  & 86.9  &89.2& 93.4&83.8\\ \hline
DFL-CNN \cite{DFL_cnn_2018} & ResNet-50& 448    & 26.3M     & 87.4  &91.7& 93.1&-\\ \hline
NTS-net \cite{nts_2018} & ResNet-50 &   448  &  25.5M    &  87.5  &91.4&93.9&-\\ \hline
TASN \cite{zheng2019looking} & ResNet-50 &   448  &  35.2M    &  87.6 &-& 93.8&-\\ \hline
\hline
Ours (single scale) & ResNet-50& 448& 23.9M & 87.7 & 92.1 & 94.3 & 87.1\\ \hline
Ours (multi scale) & ResNet-50& 448& 23.9M & \textbf{88.6}&\textbf{93.5}&\textbf{94.9}&\textbf{87.7}\\ \hline
\end{tabular}}
\caption{Experimental results on four standard datasets. ``-'' means the information is not mentioned in the relevant paper. 
 Our method outperforms existing approaches on four commonly used fine-grained datasets, and  requires no additional parameters compared to the ResNet-50 backbone.
Here, the parameters are computed on CUB-200-2011, having 200 output classes.}
\label{tab:cub}
\end{table*}

We observe that our method achieves the best accuracy on birds classification task (Table \ref{tab:cub}). Specifically, our method obtains an accuracy of 88.6\% which outperforms TASN (87.6\%) \cite{zheng2019looking}. TASN \cite{zheng2019looking} performs well because it first uses a small network to find the attentive regions and then distills knowledge from various informational regions to the model. With a low parametric complexity, our method can capture more relevant regions by focusing on hard classes and diversifying informative areas in the class activation maps.

For other four datasets, our method also outperforms the compared methods. In Aircraft, we achieve 93.5\% top-1 accuracy, surpassing NTS-net \cite{nts_2018} (91.4\%). In Cars, we obtain 94.9\%, outperforming the best performances: 93.8\% of TASN \cite{zheng2019looking}. In Dogs, we obtain 87.7\% top-1 accuracy compared to 87.3\% obtained by RACNN approach \cite{racnn_2017}. Note that RACNN has much more parameters (429M) than our methods (23.9M). 
In GTOS-mobile, we show our result using ResNet-50 with "single scale", for fair comparison with  Deep-TEN \cite{zhang2017deep} and DEP \cite{dataset_gtosmobile}. We get 85.0\%, which is 2.8\% better than the current state-of-the-art 82.2\%.

\begin{table}[h]
\centering
\resizebox{\columnwidth}{!}{
\begin{tabular}{|c|c|}
\hline
Methods        & Accuracy\\ \hline
B-CNN \cite{lin2015bilinear}&75.4  \\\hline
Deep-TEN \cite{zhang2017deep} &  76.1 \\\hline
DEP \cite{dataset_gtosmobile}  & 82.2\\\hline \hline
Ours (single scale) & \textbf{85.0}\\ \hline
\end{tabular}}
\caption{Experimental results on GTOS-mobile.}
\label{tab:gtos}
\end{table}

\subsection{Ablation Study}
To fully analyze our method, Table \ref{tab:ablation_study} provides a detailed ablation analysis on the key components of our method. It basically highlights the importance of diversification block and gradient-boosting loss. We conduct all ablation studies on CUB-200-2011 using the ResNet-50 \cite{resnet_2016}.

\textbf{Diversification block (DB)}. DB is important because it diversifies the informative regions by forcing the network to find relevant parts other than the most obvious ones. Integrating DB block in the ResNet-50 backbone results in a performance improvement of 0.8\%  (from 85.5\% to 86.3\%). 

\begin{table}[h]
\centering
\resizebox{\columnwidth}{!}{
\begin{tabular}{|c|c|}
\hline
Methods  & Accuracy \\ \hline \hline
ResNet-50  &   85.5   \\ \hline 
ResNet-50+DB  &  86.3    \\ \hline
ResNet-50+DB+Center loss \cite{center_loss}  &   86.4   \\ \hline
ResNet-50+DB+LGM loss \cite{lgm_loss} &   86.5  \\ \hline
ResNet-50+DB+MaxEnt \cite{max_entropy} &  86.2    \\ \hline
Ours (single scale)    &  \textbf{87.7} \\ \hline
Ours (multi scale)    &  \textbf{88.6} \\ \hline

\end{tabular}}
\caption{Ablation analysis on the CUB-200-2011. Our diversification block (DB) and gradient-boosting loss provide progressive improvements over the baseline.}
\label{tab:ablation_study}
\end{table}

\textbf{Gradient-boosting loss}. Our gradient-boosting loss is another important component that shows significant improvement. Using this loss, we improve the results from 86.3\% to 87.7\% (an absolute gain of 1.4\%).
 We also compare our loss with other recent losses that aim at achieving better discriminability: Center loss \cite{center_loss}, LGM loss \cite{lgm_loss}, and max entropy loss \cite{max_entropy}.  The results show that gradient-boosting loss outperforms all these loss functions. Our loss targets on difficult/confusing classes and selectively boosts the gradients for them, while other losses consider all negative classes as equal.
\begin{table}[t]
\centering
\begin{tabular}{| *{5}{c|}}
\hline
\centering
   $\alpha$   & 0 & 0.1 &0.2 & 1.0 \\ \hline
Accuracy & 86.2  &  \textbf{86.3}   &  85.8   & 85.5 \\ \hline
\end{tabular}
\caption{Ablation study on suppressing factor $\alpha$. Keeping suppressing factor as small leads to good performance.}
\label{tab:ablation_alpha}
\end{table}


\begin{figure}[t]
		\centering
		\includegraphics[width=0.6\columnwidth]{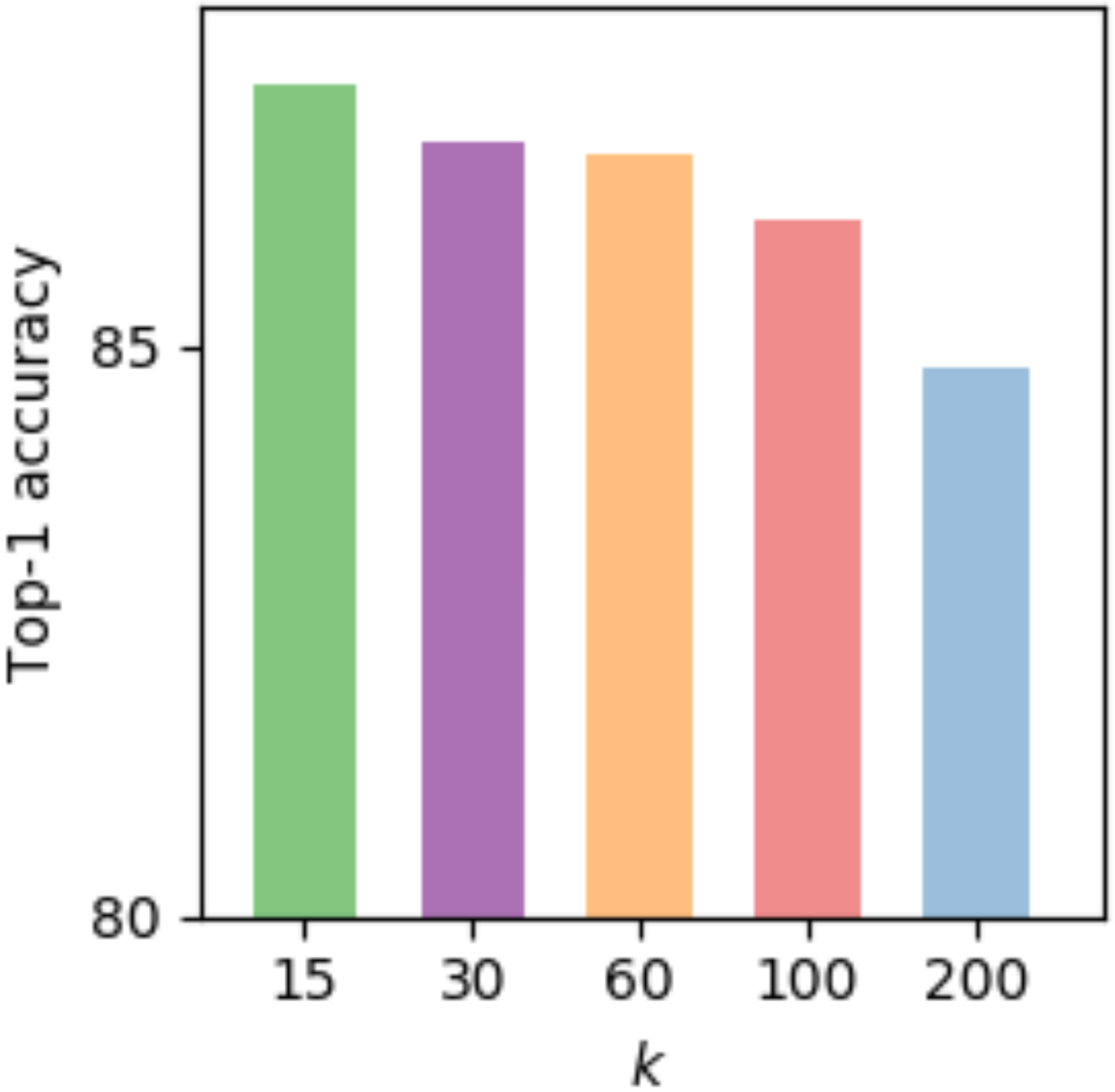}
	\caption{Ablation study on $k$ for our loss in CUB-200-2011.}
	\label{fig:loss_k}
\end{figure}

\textbf{Suppressing Factor}.
Here, we show a parameter sensitivity analysis on the suppressing factor $\alpha$. Top-1 accuracy with respect to different $\alpha$ settings is shown in Table~\ref{tab:ablation_alpha}. It shows that keeping $\alpha$ as a small value consistently leads to better performance than without using diversification block ($\alpha=1$). Specifically, $\alpha=0.1$ gives the best performance on CUB-200-2011 dataset.
\begin{figure*}[t]
\centering
\includegraphics[width=0.95\linewidth]{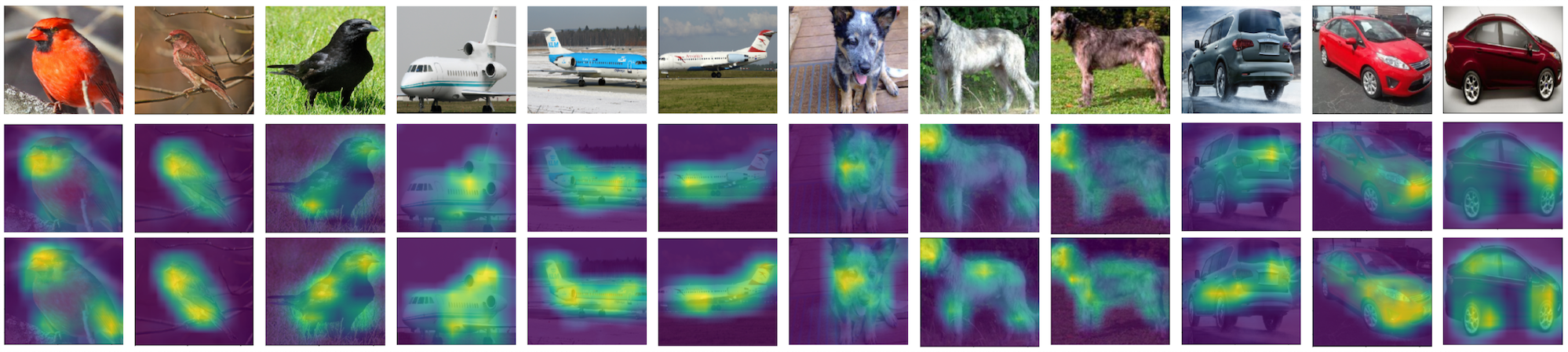}\\
\caption{Class activation map (CAM) comparison between our method and baseline in different datasets. \emph{Top} to \emph{below}: original image, CAM of the ground-truth class of baseline, CAM of the ground-truth class of our method. While baseline only focuses on the most discriminative region, our method accurately diversifies attentions to other informative regions of the objects. }
\label{Fig:qual_result}
\end{figure*}
\begin{figure}[htp]
\centering
\includegraphics[width=0.86\columnwidth]{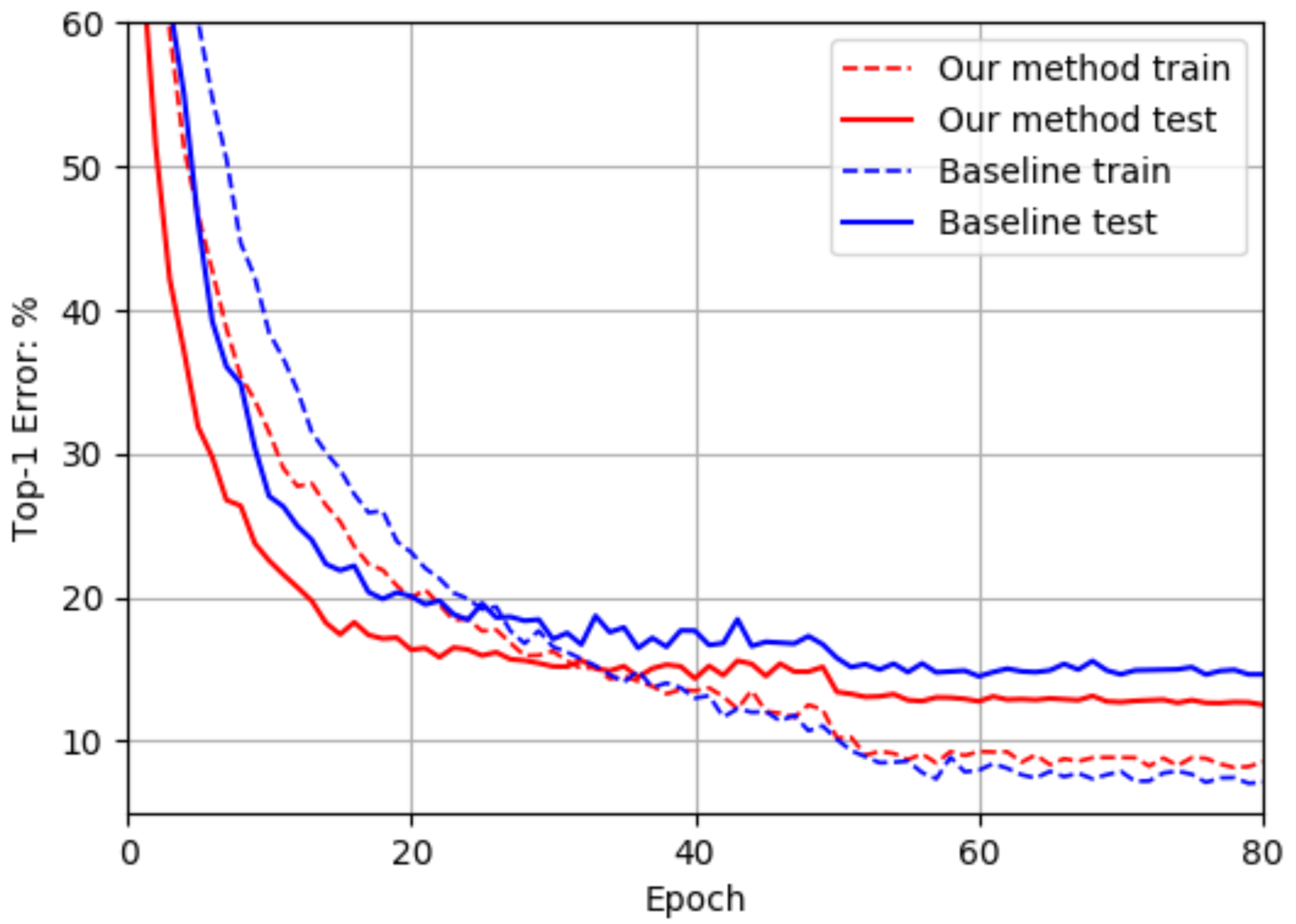}\\
\caption{Training curves of our methods and the baseline (CE loss) on CUB-200-2011. Using our loss, our method converges faster and performs better than the baseline.}
	\label{fig:converge}
\end{figure}
\begin{figure}[t]
\centering
\includegraphics[width=0.89\columnwidth]{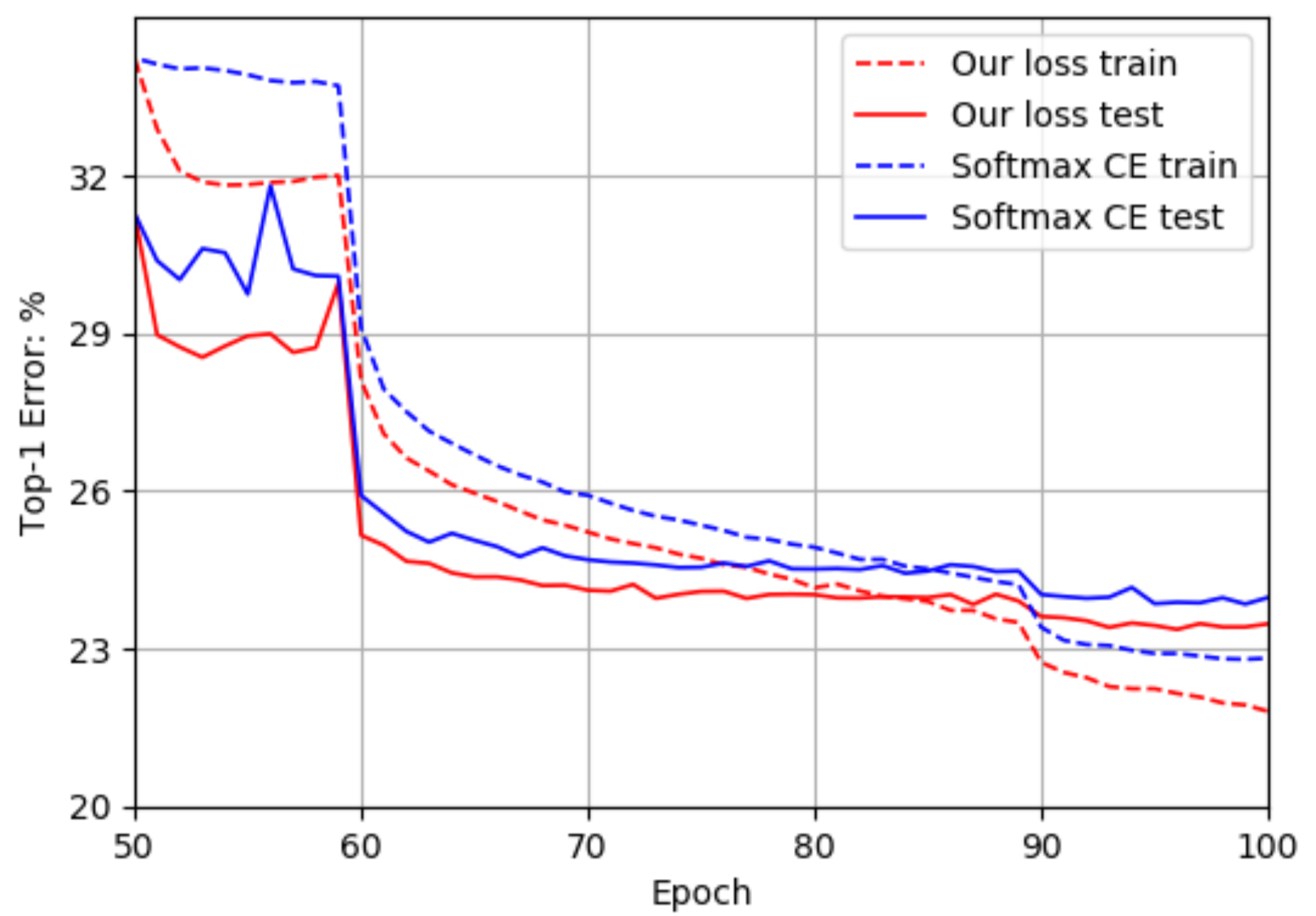}\\
	\caption{Training curves of using our loss and Cross Entropy (CE) on ImageNet. Our loss converges considerably faster than CE and also leads to a better performance. 
	}
	\label{fig:imagenet}
\end{figure}

\textbf{Choices of $k$}.
Here, we show ablation study on $k$, the number of negative classes to focus on for gradient-boosting loss, in Fig. \ref{fig:loss_k}. It shows that by reducing $k$, our loss focuses on more confusing classes and achieves consistent improvement in top-1 accuracy. 

\textbf{Convergence Analysis}.
We compare the training curves of our methods and baseline (ResNet-50) in Fig.~\ref{fig:converge}. It shows that our method converges much faster than the baseline, and also attains a lower error rate on test set. Remarkably, the baseline achieves a lower error rate on training set after 50 epochs, but fails to generalize well to the test set. This shows that the baseline is prone to overfitting on the train set, which our method successfully avoids. Our method uses diversification block which prevents the network to only focus on the most discriminative regions, like the beak or head of a bird. In contrast, using diversification block, the network finds various informative areas, thus reducing overfitting.

\subsection{Qualitative Results}
We qualitatively illustrate the comparison between our approach and the baseline in Fig.~\ref{Fig:qual_result}. We note that the diversification block indeed helps the network to find more discriminative regions in the image. In contrast, the baseline model generally focuses on the most obvious distinguishing patterns and its attention is limited to only a limited set of spatial locations. This explains why our approach generalizes better to test images, attaining a higher accuracy (Fig.~\ref{fig:converge}).

\subsection{ImageNet Results}
To validate the generality of gradient-boosting loss in visual recognition, we apply it to a ResNet-50 backbone on ImageNet \cite{ILSVRC15}. Here, we use the input size of 224$\times$224 and follow the same training strategy as used in \cite{resnet_2016}. Since our loss focuses on difficult classes, we apply it only half way (50 epochs) during training when easy categories are already well-classified. The comparison of training curve between using our loss and using cross entropy (CE) is shown in Fig. \ref{fig:imagenet}. It shows that the proposed loss converges much faster than the CE loss and achieves a lower error rate on the challenging ImageNet benchmark. 

\section{Conclusion}
We proposed a novel approach to better discriminate closely related categories in fine-grained classification task. Our method has two novel components: (a) diversification block that forces the network to find subtle distinguishing features between each pair of classes and (b) gradient-boosting loss that specifically focuses on maximally separating the highly similar and confusing classes. Our approach not only outperforms existing methods on all studied fine-grained datasets, but also demonstrates much faster convergence rates. In comparison to previous methods, our solution is both simple and elegant, leads to higher accuracy and demonstrates better computational efficiency. 

\bibliographystyle{aaai}
\bibliography{egbib}

\begin{thebibliography}{}

\bibitem[\protect\citeauthoryear{Berg and
  Belhumeur}{2013}]{fine_grain_annotation1}
Berg, T., and Belhumeur, P.
\newblock 2013.
\newblock Poof: Part-based one-vs.-one features for fine-grained
  categorization, face verification, and attribute estimation.
\newblock In {\em CVPR}.

\bibitem[\protect\citeauthoryear{Chen \bgroup et al\mbox.\egroup
  }{2019}]{Chen_2019_CVPR}
Chen, Y.; Bai, Y.; Zhang, W.; and Mei, T.
\newblock 2019.
\newblock Destruction and construction learning for fine-grained image
  recognition.
\newblock In {\em CVPR}.

\bibitem[\protect\citeauthoryear{Chhavi}{2018}]{max_heap}
Chhavi.
\newblock 2018.
\newblock k largest(or smallest) elements in an array-added min heap method.

\bibitem[\protect\citeauthoryear{Cui \bgroup et al\mbox.\egroup
  }{2018}]{cui2018large}
Cui, Y.; Song, Y.; Sun, C.; Howard, A.; and Belongie, S.
\newblock 2018.
\newblock Large scale fine-grained categorization and domain-specific transfer
  learning.
\newblock In {\em CVPR}.

\bibitem[\protect\citeauthoryear{Dubey \bgroup et al\mbox.\egroup
  }{2018a}]{pc_2018}
Dubey, A.; Gupta, O.; Guo, P.; Raskar, R.; Farrell, R.; and Naik, N.
\newblock 2018a.
\newblock Pairwise confusion for fine-grained visual classification.
\newblock In {\em ECCV}.

\bibitem[\protect\citeauthoryear{Dubey \bgroup et al\mbox.\egroup
  }{2018b}]{max_entropy}
Dubey, A.; Gupta, O.; Raskar, R.; and Naik, N.
\newblock 2018b.
\newblock Maximum-entropy fine grained classification.
\newblock In {\em NIPS}.

\bibitem[\protect\citeauthoryear{Fu, Zheng, and Mei}{2017}]{racnn_2017}
Fu, J.; Zheng, H.; and Mei, T.
\newblock 2017.
\newblock Look closer to see better: Recurrent attention convolutional neural
  network for fine-grained image recognition.
\newblock In {\em CVPR}.

\bibitem[\protect\citeauthoryear{Gao \bgroup et al\mbox.\egroup }{2016}]{cbp}
Gao, Y.; Beijbom, O.; Zhang, N.; and Darrell, T.
\newblock 2016.
\newblock Compact bilinear pooling.
\newblock In {\em CVPR}.

\bibitem[\protect\citeauthoryear{Ge, Lin, and Yu}{2019}]{ge2019weakly}
Ge, W.; Lin, X.; and Yu, Y.
\newblock 2019.
\newblock Weakly supervised complementary parts models for fine-grained image
  classification from the bottom up.
\newblock In {\em CVPR}.

\bibitem[\protect\citeauthoryear{He \bgroup et al\mbox.\egroup
  }{2016}]{resnet_2016}
He, K.; Zhang, X.; Ren, S.; and Sun, J.
\newblock 2016.
\newblock Deep residual learning for image recognition.
\newblock In {\em CVPR}.

\bibitem[\protect\citeauthoryear{He \bgroup et al\mbox.\egroup
  }{2017}]{maskrcnn}
He, K.; Gkioxari, G.; Doll{\'a}r, P.; and Girshick, R.~B.
\newblock 2017.
\newblock Mask r-cnn.
\newblock {\em ICCV}.

\bibitem[\protect\citeauthoryear{Khosla \bgroup et al\mbox.\egroup
  }{2011}]{dataset_dogs}
Khosla, A.; Jayadevaprakash, N.; Yao, B.; and Fei-Fei, L.
\newblock 2011.
\newblock Novel dataset for fine-grained image categorization.
\newblock In {\em CVPR Workshop}.

\bibitem[\protect\citeauthoryear{Krause \bgroup et al\mbox.\egroup
  }{2013}]{dataset_cars}
Krause, J.; Stark, M.; Deng, J.; and Fei-Fei, L.
\newblock 2013.
\newblock 3d object representations for fine-grained categorization.
\newblock In {\em 4th International IEEE Workshop on 3D Representation and
  Recognition}.

\bibitem[\protect\citeauthoryear{Li \bgroup et al\mbox.\egroup
  }{2017}]{li2017dynamic}
Li, Z.; Yang, Y.; Liu, X.; Zhou, F.; Wen, S.; and Xu, W.
\newblock 2017.
\newblock Dynamic computational time for visual attention.
\newblock In {\em ICCV}.

\bibitem[\protect\citeauthoryear{Li \bgroup et al\mbox.\egroup
  }{2018}]{Li_2018_CVPR}
Li, P.; Xie, J.; Wang, Q.; and Gao, Z.
\newblock 2018.
\newblock Towards faster training of global covariance pooling networks by
  iterative matrix square root normalization.
\newblock In {\em CVPR}.

\bibitem[\protect\citeauthoryear{Lin \bgroup et al\mbox.\egroup
  }{2014}]{coco_eccv2014}
Lin, T.-Y.; Maire, M.; Belongie, S.; Hays, J.; Perona, P.; Ramanan, D.;
  Doll{\'a}r, P.; and Zitnick, C.~L.
\newblock 2014.
\newblock Microsoft coco: Common objects in context.
\newblock In {\em ECCV}.

\bibitem[\protect\citeauthoryear{Lin, RoyChowdhury, and
  Maji}{2015}]{lin2015bilinear}
Lin, T.-Y.; RoyChowdhury, A.; and Maji, S.
\newblock 2015.
\newblock Bilinear cnn models for fine-grained visual recognition.
\newblock In {\em ICCV}.

\bibitem[\protect\citeauthoryear{Maji \bgroup et al\mbox.\egroup
  }{2013}]{dataset_aircraft}
Maji, S.; Kannala, J.; Rahtu, E.; Blaschko, M.; and Vedaldi, A.
\newblock 2013.
\newblock Fine-grained visual classification of aircraft.
\newblock Technical report.

\bibitem[\protect\citeauthoryear{Paszke \bgroup et al\mbox.\egroup
  }{2017}]{paszke2017automatic}
Paszke, A.; Gross, S.; Chintala, S.; Chanan, G.; Yang, E.; DeVito, Z.; Lin, Z.;
  Desmaison, A.; Antiga, L.; and Lerer, A.
\newblock 2017.
\newblock Automatic differentiation in pytorch.
\newblock In {\em NIPS Workshop}.

\bibitem[\protect\citeauthoryear{Qian \bgroup et al\mbox.\egroup
  }{2015}]{qian2015fine}
Qian, Q.; Jin, R.; Zhu, S.; and Lin, Y.
\newblock 2015.
\newblock Fine-grained visual categorization via multi-stage metric learning.
\newblock In {\em CVPR}.

\bibitem[\protect\citeauthoryear{Russakovsky \bgroup et al\mbox.\egroup
  }{2015}]{ILSVRC15}
Russakovsky, O.; Deng, J.; Su, H.; Krause, J.; Satheesh, S.; Ma, S.; Huang, Z.;
  Karpathy, A.; Khosla, A.; Bernstein, M.; Berg, A.~C.; and Fei-Fei, L.
\newblock 2015.
\newblock {ImageNet Large Scale Visual Recognition Challenge}.
\newblock {\em IJCV}.

\bibitem[\protect\citeauthoryear{Singh and Lee}{2017}]{has_2017}
Singh, K.~K., and Lee, Y.~J.
\newblock 2017.
\newblock Hide-and-seek: Forcing a network to be meticulous for
  weakly-supervised object and action localization.
\newblock In {\em ICCV}.

\bibitem[\protect\citeauthoryear{Sun \bgroup et al\mbox.\egroup
  }{2018}]{mamc_2018}
Sun, M.; Yuan, Y.; Zhou, F.; and Ding, E.
\newblock 2018.
\newblock Multi-attention multi-class constraint for fine-grained image
  recognition.
\newblock In {\em ECCV}.

\bibitem[\protect\citeauthoryear{Szegedy \bgroup et al\mbox.\egroup
  }{2016}]{szegedy2016rethinking}
Szegedy, C.; Vanhoucke, V.; Ioffe, S.; Shlens, J.; and Wojna, Z.
\newblock 2016.
\newblock Rethinking the inception architecture for computer vision.
\newblock In {\em CVPR}.

\bibitem[\protect\citeauthoryear{Van Der~Maaten}{2014}]{van2014accelerating}
Van Der~Maaten, L.
\newblock 2014.
\newblock Accelerating t-sne using tree-based algorithms.
\newblock {\em JMLR}.

\bibitem[\protect\citeauthoryear{Van~Horn \bgroup et al\mbox.\egroup
  }{2018}]{van2018inaturalist}
Van~Horn, G.; Mac~Aodha, O.; Song, Y.; Cui, Y.; Sun, C.; Shepard, A.; Adam, H.;
  Perona, P.; and Belongie, S.
\newblock 2018.
\newblock The inaturalist species classification and detection dataset.
\newblock In {\em CVPR}.

\bibitem[\protect\citeauthoryear{Wah \bgroup et al\mbox.\egroup
  }{2011}]{dataset_cub}
Wah, C.; Branson, S.; Welinder, P.; Perona, P.; and Belongie, S.
\newblock 2011.
\newblock The caltech-ucsd birds-200-2011 dataset.

\bibitem[\protect\citeauthoryear{Wan \bgroup et al\mbox.\egroup
  }{2018}]{lgm_loss}
Wan, W.; Zhong, Y.; Li, T.; and Chen, J.
\newblock 2018.
\newblock Rethinking feature distribution for loss functions in image
  classification.
\newblock In {\em CVPR}.

\bibitem[\protect\citeauthoryear{Wang \bgroup et al\mbox.\egroup
  }{2016}]{wang2016mining}
Wang, Y.; Choi, J.; Morariu, V.; and Davis, L.~S.
\newblock 2016.
\newblock Mining discriminative triplets of patches for fine-grained
  classification.
\newblock In {\em CVPR}.

\bibitem[\protect\citeauthoryear{Wang, Morariu, and Davis}{2018}]{DFL_cnn_2018}
Wang, Y.; Morariu, V.~I.; and Davis, L.~S.
\newblock 2018.
\newblock Learning a discriminative filter bank within a cnn for fine-grained
  recognition.
\newblock In {\em CVPR}.

\bibitem[\protect\citeauthoryear{Wen \bgroup et al\mbox.\egroup
  }{2016}]{center_loss}
Wen, Y.; Zhang, K.; Li, Z.; and Qiao, Y.
\newblock 2016.
\newblock A discriminative feature learning approach for deep face recognition.
\newblock In {\em ECCV}.

\bibitem[\protect\citeauthoryear{Xue \bgroup et al\mbox.\egroup
  }{2017}]{dataset_gtos}
Xue, J.; Zhang, H.; Dana, K.; and Nishino, K.
\newblock 2017.
\newblock Differential angular imaging for material recognition.
\newblock In {\em CVPR}.

\bibitem[\protect\citeauthoryear{Xue, Zhang, and
  Dana}{2018}]{dataset_gtosmobile}
Xue, J.; Zhang, H.; and Dana, K.
\newblock 2018.
\newblock Deep texture manifold for ground terrain recognition.
\newblock In {\em CVPR}.

\bibitem[\protect\citeauthoryear{Yang \bgroup et al\mbox.\egroup
  }{2018}]{nts_2018}
Yang, Z.; Luo, T.; Wang, D.; Hu, Z.; Gao, J.; and Wang, L.
\newblock 2018.
\newblock Learning to navigate for fine-grained classification.
\newblock In {\em ECCV}.

\bibitem[\protect\citeauthoryear{Zhang, Xue, and Dana}{2017}]{zhang2017deep}
Zhang, H.; Xue, J.; and Dana, K.
\newblock 2017.
\newblock Deep ten: Texture encoding network.
\newblock In {\em CVPR}.

\bibitem[\protect\citeauthoryear{Zheng \bgroup et al\mbox.\egroup
  }{2017}]{zheng2017learning}
Zheng, H.; Fu, J.; Mei, T.; and Luo, J.
\newblock 2017.
\newblock Learning multi-attention convolutional neural network for
  fine-grained image recognition.
\newblock In {\em ICCV}.

\bibitem[\protect\citeauthoryear{Zheng \bgroup et al\mbox.\egroup
  }{2019}]{zheng2019looking}
Zheng, H.; Fu, J.; Zha, Z.-J.; and Luo, J.
\newblock 2019.
\newblock Looking for the devil in the details: Learning trilinear attention
  sampling network for fine-grained image recognition.
\newblock In {\em CVPR}.

\bibitem[\protect\citeauthoryear{Zhou \bgroup et al\mbox.\egroup }{2016}]{CAMs}
Zhou, B.; Khosla, A.; Lapedriza, A.; Oliva, A.; and Torralba, A.
\newblock 2016.
\newblock Learning deep features for discriminative localization.
\newblock In {\em CVPR}.

\end{thebibliography}

\end{document}